# What Makes a Message Persuasive?
## Identifying Adaptations Towards Persuasiveness in Nine Exploratory Case Studies

Sebastian Duerr[1], Krystian Teodor Lange[2], Peter A. Gloor[1]
[1]MIT Center for Collective Intelligence, [2]MPP Candidate, Harvard Kennedy School
contact: {sduerr, pgloor}@mit.edu, tlange@hks.harvard.edu

The ability to persuade others is critical to professional and personal success. However, crafting persuasive messages is demanding and poses various challenges. We conducted nine exploratory case studies to identify adaptations that professional and non-professional writers make in written scenarios to increase their subjective persuasiveness. Furthermore, we identified challenges that those writers faced and identified strategies to resolve them with persuasive natural language generation, i.e., artificial intelligence. Our findings show that humans can achieve high degrees of persuasiveness (more so for professional-level writers), and artificial intelligence can complement them to achieve increased celerity and alignment in the process.

## 1. Introduction

A U.S.-focused survey found that people spend 3.1 hours per weekday on average on their work email and an additional 2.5 hours on their personal email. Assuming an average 8-hour workday, employees spend more than one third of their time communicating through writing (Adobe 2018). These numbers were compiled before COVID-19 (Liu et al. 2020). They indicate that the time spent on email is substantial and may therefore be optimized.

In this research, we identify multiple reasons why drafting and rewriting emails can take a significant amount of time. For instance, finding the right tone or conveying the correct level of politeness and formality made some persuaders in our experiments go back and forth multiple times to increase the persuasiveness of their writing. In many situations, a persuasive Natural Language Generation (NLG) artificial intelligence (AI) can help resolve such uncertainties that arise in writing persuasive messages (Duerr & Gloor 2021).

A persuasive NLG AI is a system that can create communications aimed at a user (the persuadee or recipient) to persuade them to accept a specific argument through a persuasive message (Hunter et al. 2020). With recent advances in NLP and its subfield of Natural Language Generation (NLG), it was demonstrated that pre-trained language models (e.g., GPT-3) can achieve state-of-the-art results on a wide range of NLP tasks (Economist 2020). Such models allow for writing human-like texts through NLG and can be utilized to engage in computational persuasion (McNamara et al. 2013, Tan et al. 2016, Zarouali et al. 2020, Douven & Mirabile 2018, Hunter et al. 2020).

In their literature review, Duerr & Gloor (2021) propose how working with the recipient's subsequent uptake (Iyer & Sycara 2019) can be achieved through determinants that may constitute a persuasive NLG. The authors conclude a research agenda with the proposition to investigate: 'How [...] successful persuasiveness in Natural Language Generation [should] be theorized [...]' (Duerr & Gloor 2021, p. 11). Following this proposition, we formulate our research question as:

*How does the interplay of human and AI relate to persuasion success within a written message context?*

Methodologically, we conducted exploratory case studies with four individuals with professional-level persuasive writing skills and five writers with non-professional persuasive writing skills. These individuals were selected from five different cultures to derive, categorize, and differentiate between persuasive adaptations. The findings were aligned to identify complementarities among human and artificial intelligence adaptations towards increased pervasiveness. Theoretically, we build on Aristotle's rhetorical ontology (ethos, logos, pathos, and kairos; Schiappa & Nordin 2013) and use it as a categorical framework.

The following sections review the literature on theories related to persuasion and Aristotle's ontology, introduce and analyze the exploratory case studies, and then discuss the main findings, their limitations, and implications for future research.

## 2. Theoretical Conceptualization of Persuasiveness

### 2.1 Artificial and Human Intelligence Towards Persuasiveness

*Artificial intelligence towards persuasiveness* pertains to techniques in NLP or NLG that aim to increase a text's persuasiveness (Duerr & Gloor 2021). With computing integrating into every dimension of life, persuasion becomes a target for applying computer-based approaches (Zarouali et al. 2020).

Related research on persuasion is conducted by Guerini et al. (2008). The authors analyze nine hundred tagged speeches on a lexical level and extract "persuasive words." They hypothesize that tags about audience reaction (e.g., applause) are hot-spots indicators where persuasion attempts succeed. In a different experiment, Iyer and Sycara (2019) analyze political speeches and use machine learning to classify transcripts of political discourse according to their persuasive power and tactics. Next, Atalay et al. (2019), in 'Using NLP to Investigate Syntactic Structure,' investigate the impact of language syntax and readability on a communicated message's persuasiveness. The authors find that persuasive tactics have inherent sentential structures. Tan et al. (2016) look at tweets' wordings to determine popularity in opposition to author or topic popularity. As a finding, they prove their methods to perform better than humans do on average. In addition to textual persuasion scenarios, there has also been research conducted on persuasion in multimedia (e.g., Siddiquie et al. 2015, Chatterjee et al. 2014, Lin et al. 2006, Park et al. 2014, Bikeland et al. 2007, Iyer et al. 2019). However, we do not consider this stream for the presented research on NLP since we limit our research engagement to written persuasion. As a delineation from previous research, our focus on explicit transformations towards increased persuasiveness is not thoroughly explored in research yet (Iyer & Sycara 2019).

*Human intelligence towards persuasiveness* pertains to situations in which the *persuader* imposes a mental state in the recipient, for instance, through praise or threats. Still, in contrast to a regular expression of sentiment, a persuasive act intends to change the recipient's beliefs (Iyer & Sycara 2019). In our work, we define a persuader based on the work of contemporary communication and psychological research (Rocklage et al. 2018, Park et al. 2015, Hunter et al. 2019) as the persuader as acting intentionally.

In research, we identified several ontologies (e.g., Cialdini & Goldstein 2002, Quirk et al. 1985, Brewer 1980). Yet, we will advance with the most established from around 350 BCE by Aristotle (Schiappa & Nordin 2013) and reflect it from the perspective of a contemporary business framework of persuasion (DeMers 2016).

## 2.2. Aristotle Categories on Persuasiveness

In his seminal work '*On Rhetoric'*, Aristotle introduced his widely-known ontology for persuasive acts. Accordingly, persuasion depends on multiple facets: appeal to emotions of the recipient (*pathos*), logical structure of the argument (*logos*), on the character of the speaker themselves (*ethos*), and the contextual timing (*kairos*) (Schiappa & Nordin 2013). Likewise, contemporary business literature conceptualizes persuasive acts through "principles of persuasion" (DeMers 2016) which we consolidate alongside Aristotle's categories:

Arguments with consistent *logic* in persuasive acts increase persuasiveness (Cialdini & Goldstein 2002, Walton et al. 2008, Block et al. 2019). This claim is in line with the theory of probabilistic models (McGuire 1981, Wyer 1970). It is assumed that conclusive statements lead to a recipient's expectation that a conclusion will follow. Technical implementations of logical argumentation or logical meaning representations occur as first-order logic Moens 2018) or semantic argumentation graphs (Block et al. 2019).

As in the theory of cognitive dissonance (Festinger 1957), *benevolence* aims to create value for the recipient (DeMers 2016, Voss & Raz 2016). Persuasive manipulations relate to altering a recipient's perceived benevolence through dissonant or consonant measures. An implementation in a persuasive NLG AI can be facilitated by identifying their absence or impact (Hunter et al. 2019, Zarouali et al. 2020).

*Linguistic appropriacy* subsumes adaptations that facilitate an individual's stylome. Such a stylome can be quantified and identified linguistically (Zarouali et al. 2020). Language expectation theory identifies written or spoken language as a rule-based system through which recipients develop expectations (Burgoon & Miller 1985). The reason for profiling the stylome of an individual is to match their expectations (Park et al. 2015). Once implemented, a persuasive NLG AI can achieve congruence between a persuasive message and the recipient, and thus, create persuasiveness (Duerr & Gloor 2021).

*Trust* plays an essential role in the persuader-recipient relationship. If established, the recipient's attitude toward the persuader - as identified in balance theory (Heider 1958) - helps a recipient to reason about the reciprocating nature, honesty, or reliability of the counterpart (Kim & Duhachek 2020). An implementation of trustworthiness can, amongst others, be realized through identifying a recipient's psychological profile to influence the degree of persuasiveness of a persuasive act (e.g., extroverted individuals respond better to messages that have a positive valence and that are, in that case, more persuadable, Zarouali et al. 2020, Park et al. 2015).

After the description of the method that we used to conduct our exploratory case studies in section 3, section 4 will utilize the above-presented categories of Aristotle to classify the adaptations that humans make to increase the persuasiveness of a text.

## 3. Method

We conducted nine exploratory interviews to understand how individual writers approach the task of increasing the persuasiveness of a given email by changing it. Based on the recommendations of Eisenhardt (1989) and Yin (2009), the interviews followed semi-structured guidelines with open-ended questions to assure the examination of possible research direction. We conducted the interviews with both individuals with professional-level persuasive writing skills and those without. The individuals came from various backgrounds such as diplomacy, literature, or marketing. The interviewees were not limited to geographical regions to learn about differences in cultures. The interviews took between 40 and 90 minutes and were conducted in English. Table 1 provides an overview of the cultural background, persuasive writing levels, the interviewees' professional background, the length of the interview in minutes, and time spent on the persuasion exercise:

| Cultural Background, Gender | Persuasive Writing Level | ID | Background | Interview Length [min] | Persuasion Exercise [min] |
|---|---|---|---|---|---|
| Australian, m | Professional level | IP1 | Diplomat | 37 | 19 |
| American, m | Professional level | IP2 | Policy Maker in Education | 67 | 25 |
| American, f | Non-professional level | IP3 | Marketing Manager | 62 | 28 |
| American, m | Professional level | IP4 | Graduate Student in Digital Government | 52 | 12 |
| British, f | Non-professional level | IP5 | Technical Support Engineer | 64 | 30 |
| British, m | Professional level | IP6 | Private Investigator | 52 | 20 |
| Continental European, f | Non-professional level | IP7 | Graduate Student in Mgmt. & Technology | 97 | 32 |
| Continental European, m | Non-professional level | IP8 | Graduate Student in Medicine | 75 | 34 |
| Middle East, m | Non-professional level | IP9 | Medical Science Liaison | 39 | 29 |

Table 1.Culture, Language Level, and Interviewees

Interviews took place virtually through the video-chat tool Zoom (version 5.0.3) and were recorded with the consent of the interview partners. Normally, two of the authors conducted the sessions. Our interview guideline is designed as follows: First, the participants were enquired about their professional background and their experience regarding persuasion. Next, an interviewee was provided with an email draft (two different American business scenarios, i.e., applying for a loan and asking for a promotion). They were asked to adapt so that the text would become more persuasive, i.e. creating a way to a positive conversation on the persuader's ask. Participants were asked to think aloud and justify their adaptations towards persuasiveness while sharing their screen with the interviewers. In the third part of the interview, participants were asked about writing persuasive messages and questions about data privacy, algorithm trust, and AI-powered assistants.

The interviews took place from February 2021 to March 2021. All interviews were transcribed, and the documentations, profiles, off-record notes, and observations were used to augment and triangulate the interview data. Data analysis progressed gradually through iterations between coding, looking for meanings in the data, writing narratives, and revisiting literature. Following Miles and Huberman's (1994) recommendations, this data analysis process was facilitated through the building of data displays in the form of tables and matrices (through coding in MaxQDA2020 v. 20.4.0) to refine the concepts identified and the development of tentative conclusions to depict the emerging adaptations (Duerr et al. 2017).

The data analysis began with descriptive codes (or open coding) as soon as the first interviews were transcribed and were done inductively, seeking to reflect the data as closely as possible. This stage led to the identification of over 140 descriptive coded statements. The focus was on interpreting the data to search for mechanisms, relationships, and patterns and facilitate the next stage, pattern coding. The process was highly iterative, moving between description and interpretation and between data and theory. As our output, we classified the mechanisms into an ontology of four categories by Aristotle (Schiappa & Nordin 2013). Recognizing the classification, we looked back at the data and focused on coding around these four concepts, which are presented in the next section. Finally, we aligned human adaptations with the findings of the literature review by Duerr & Gloor (2021) to identify which determinants artificial intelligence can contribute to and which complementarities evolve from their use.

# 4. Results

## 4.1 Human Adaptations to Increase Persuasiveness of Text

Human adaptations to increase the persuasiveness of texts are organized alongside the Aristotle categories (Table 2, column 1). We identified them by observing the adaptations of the interview partners through the recording while marking those instances in the respective transcript. Table 2 lists the adaptations that our interviewees used (column 2) and related quotes in which we identified those adaptations (column 3).

| Category | Human Adaptations | Related Quotes |
|---|---|---|
| Language | <ul><li>Identifies changes in tone, and corrects tone based on the authority of the recipient.*</li><li>Chooses different wording strategies.</li></ul> | <ul><li>The door isn't open for a conversation, I'll change that. (IP2)</li><li>I'll put the 'dear' to make it more formal. (IP1)</li><li>I guess any other word there would work better like recently or this year. (IP3)</li><li>I will stick to last names although she's not. I don't want to be [...] construed as rude depending on what her world is. (IP5)</li><li>It should be easy to read but because I'm [non-native] I had to read this one twice. (IP7)</li></ul> |
| Logic | <ul><li>Integrates logical arguments that may be only relevant to the recipient in that scenario.*</li><li>Reframes the situation through the usage of analogies or metaphors.*</li><li>Applies structural changes through order or via bullet points.</li></ul> | <ul><li>Her text is formatted indeed, but not like nicely structured. (IP3)</li><li>She wants to focus the money in her appeal. Hence, I'm trying to get some attention to this phrase. (IP7)</li><li>To strengthen my argument, I place the business plan and leverage she would provide me right here. (IP1)</li><li>Bullet points will make this more clear. (IP4)</li></ul> |
| Trust | <ul><li>Sets the tone with some positive emotion.</li><li>Creates interpersonal congruence through shadowing the recipient text.*</li><li>Establishes associations of recipient's mental states with good or bad traits.</li><li>Poses direct promises or threats.</li><li>Increases politeness.</li></ul> | <ul><li>I'd want to set the tone for anything by asking for something as positive. (IP2)</li><li>So, I shadow what they did. I'm also in a friendly mood. (IP3)</li><li>As a student of marketing, I learnt that you shall appeal to the bad traits because that's what people normally make decisions on. People think they've acted due to logic but they get hooked on mental state. (IP5)</li><li>I'll adapt the text to: With your support and the financial contribution, I am sure that my business will grow, and I can be a successful female entrepreneur. (IP6)</li><li>This does not sound friendly enough for an English email. It sounds like a person from a more direct culture has written it. (IP6)</li></ul> |
| Bene-volence | <ul><li>Gives related examples.</li><li>Offers additional items that may be important to the recipient but not to the persuader.*</li><li>References what is customary in a given situation.</li></ul> | <ul><li>I would augment the business situation by checking other bakeries around, to figure out how many people live in the area to support the growth, she was referring to. (IP6)</li><li>Maybe sometimes it's better to just be direct, so I help her by deleting all that is irrelevant. (IP1)</li><li>I would have added additional securities for the loan application but I knew it was factual information, so I did not do it. (IP2)</li><li>I am trying to draw attention away from the sick daughter because I want to keep the focus on this contribution and not make a personal thing. (IP2)</li></ul> |

Table 2. Adaptations by Interview Partners During the Interviews

We acknowledge that there were differences in adaptations made by professional level compared to non-professional level persuaders. The adaptations more dominantly made by professional level persuasion writers were marked with an asterisk (*). Quantitatively, we note that the interview partners took around 25 minutes (approx. 20 minutes for professional level / approx. 30 minutes for non-professional level) to increase the persuasiveness of the sample messages (approx. 180 words). Moreover, three participants (IP1, IP6, and IP8) produced a result that was about the same length or shorter than the original message the interview partners were asked to adapt.

## 4.2 Adaptations by an Artificial Intelligence to Increase Persuasiveness

During the exercise of increasing the persuasiveness of a text, we encountered certain challenges of our interview partners. Such challenges were identified either through the interviewee expressing them, through their significantly longer processing times, or by enquiring about certain difficult situations for the interviewee that the interviewers observed during the transformational task.

Table 3 lists these challenges (column 1) and related quotes (column 2). Furthermore, to address these challenges, we identified strategies by persuasive NLG AI in academic literature to provide remedy for these challenges (column 3).

| Challenge | Related Quote | Strategies to resolve through AI |
|---|---|---|
| Difficulties keeping messages concise. | 'My messages are always too long.' (IP9) | AI can optimize text conciseness through summarizations (Raffel et al. 2019). |
| Writing style selection based on gut feeling. | 'Shall I be very positive or not? I do not know.' (IP7) | Sentiment can be classified with AI and communicated to persuaders to provide orientation (Pang et al. 2002). |
| Insecurity on choosing degree of formality. | 'Can I say hi or shall it be dear?' (IP2) | Formality can be identified through AI and communicated to persuaders to provide orientation (Rao & Tetreault 2018). |
| Neglecting logical cohesion. | 'I usually do the quick adaptations, to be fast but compromise the logical thread.' (IP06) | AI can make logical transformations to integrate logic (Geva et al. 2019). |
| Defining psychological appropriacy. | 'Shall I write casual as the recipient did or adapt to a banker's tone? (IP6) | The recipients psychological traits can be inferred to provide orientation (Parker et al. 2015, Zarouali et al. 2020). |
| Excessive need of time for adaptations. | It takes me two to three hours but I really need 20 mins only for the subject'. (IP5) | Above listed AI based augmentations can reduce uncertainties (e.g., tone or formality) and therefore optimize for celerity in drafting a persuasive message. |

Table 3. Challenges Interview Partners Faced When Adapting Text

During the experiments, we learnt that our IPs did note the human cues in the text (there was a kind post scriptum note). This motivated them to adjust the tone but not the valence of the message.

## 4.3 Complementarities of Human and AI-based Adaptations to Increase Persuasiveness

In the data, we identified and analyzed the complementarities of human and artificial intelligence in persuasive scenarios. Likewise, the process of achieving persuasiveness of a text can be complemented with modern AI approaches:

First, persuasive NLG AI is able to quantify a recipient's valence (Pang et al. 2002), formality (Rao & Tetrault), or psychological traits (Parker et al. 2015, Zarouali et al. 2020). This work and previous research (Duerr & Gloor 2021) have shown that a quantification of a recipient's input text can provide orientation during the writing and give directions to mirror the degree of an own persuasive draft. Since this was a reason for frequent alterations to the non-professional level persuasive writers, this will increase celerity in writing persuasively.

Second, techniques of persuasive NLG AI can integrate logical cues (Geva et al. 2019), and sentential structure (Atalay et al. 2019), or even related convincing of arguments (Hunter et al. 2019). During the transformations of the texts, we observed that both non-professional and professional level writers neglected these aspects either due to optimizing for speed (for professional-level) or due to the inherent complexity of the persuasive rewriting complexity (for non-professional level).

However, we learned that AI inherently could not conceive the full context of a human relationship (Duerr & Gloor 2021) since not every interaction point may be documented (e.g., face-to-face meetings). Humans can integrate such cues into text, providing their entire perspective. Lastly, a persuasive NLG AI does not have information about additional meaningful items (i.e., benevolence) that may be crucial for a persuasive act to be successful that a persuader can provide.

Below, Table 4 contrasts the identified adaptations by humans and those concluded for artificial intelligence that can help the challenges that our interview partners faced in the manual persuasion sessions. Table 4 is not exhaustive due to limitations, mostly number and choice of test subjects.

|  | **Human Intelligence** | **Artificial Intelligence** |
| --- | --- | --- |
| **Identified Respective Adaptations** | <ul><li>Choose different wording strategies.</li><li>Integrate logical arguments that may be relevant only to the recipient in that scenario.</li><li>Reframes the situation through the usage of analogies.</li><li>Creates interpersonal congruence through shadowing the recipient text.</li><li>Establishes associations of recipient's mental states with good or bad traits.</li><li>Poses direct threats or promises.</li><li>References to what is customary in a given situation.</li></ul> | <ul><li>AI can make abstract summarizations (Raffel et al. 2019).</li><li>Sentiment can be classified and communicated to persuaders (Pang et al. 2002).</li><li>Formality can be identified and communicated to persuader (Rao & Tetreault 2018).</li><li>Logical transformations can be made to integrate logic (Geva et al. 2019).</li><li>The recipients psychological traits can be inferred (Parker et al. 2015, Zarouali et al. 2020).</li></ul> |
| **Identified Complement-aries** | <ul><li>AI is able to measure recipients in different dimensions and tailor tone to help with appropriate language.</li><li>AI is able to help humans with integrating logic, structure and the provision of additional arguments.</li><li>AI does not have the context of full relation of humans since not every interaction point may be documented (e.g., face to face meetings).</li><li>AI does not know about the limitations of benevolence (e.g., items for recipient) that a persuader can provide.</li></ul> | |

Table 4. The Complementaries of Artificial and Human Intelligence towards Persuasiveness

## 5. Discussion

Humans themselves achieve great persuasiveness (more so for professional-level), though it incurs temporal and mental exhaustion through frequent iterations (more so for non-professional-level). Through careful evaluation of nine exploratory case studies, we made some additional peculiar observations (i.e., professional level of persuasive writing, gender, and culture), leading to different approaches in how people address persuasiveness in text.

First, a significant difference in the degree of persuasion professionalism was that professional-level writers were consistent in creating persuasiveness through short phrasing, thorough emotional wording, aiming at similar formality, building structure, and logic (e.g., bullet pointing) while the non-professional-level writers were identifying such short-comings but needed either significantly more time or were not fully able to judge their alignment with the recipient. Consequently, we theorize "successful persuasiveness in Natural Language Generation" (Duerr & Gloor 2021, p. 11) as *the alignment of tonal formality, emotionality, and comprehensiveness with a recipient in a structured and logically coherent manner in a text.*

Second, we perceived differences in how female and male writers approached the task of persuasion. Research confers that women are more empathetic and careful than their male counterparts (Rueckert & Naybar 2008). Though we were confronted with a small sample (three females, six males), we observed more straightforwardness and confidence in their writing among the men. Since we cannot substantiate our assumptions, we invite research on the impact of gender on persuasiveness in message scenarios. Furthermore, we welcome any findings that uncover how AI interventions can facilitate optimizations, particularly for gender.

Third, in terms of cultural background, we learned that six out of nine writers emphasized the cultural aspect of writing in American business scenarios by trying to mimic what they believe the recipient would expect. As an example, our Middle East participant amended: *"I would append a sharp to 10 o'clock because in my culture it is important to say that you arrive sharp at the minute. However, in this scenario to an American bank, I would expect that anyway" (IP9).* Moreover, IP1 and IP6 mentioned that the tone of the texts and IP1 elaborated further: *"my wife means well but in her culture you communicate more directly. When I proof-read her important English emails, they sometimes sound quite mean to an English speaker."* We emphasize the importance of cultural cues in persuasive scenarios. As a future research avenue, we invite scholars to investigate the impact of different cultural backgrounds in persuasive text generation and their interdependencies. From a persuasive NLG perspective, it would be particularly interesting to advance the implementation of culture.

We contribute to existing research in persuasive NLG as we empirically theorize successful persuasiveness in NLG (section 5). Furthermore, we identify complementaries (section 4.3) between humans and artificial intelligence towards the persuasiveness of texts.

Our research needs to be reflected in the light of its limitations. We did not use chronological order or an order based on popularity due to our interview partners' heterogeneity (Duerr et al. 2017). Instead, we ordered the findings by forming a theoretical lens (cf. Section 2.2) and following this lens's aspects to classify the identified adaptations. In addition, there may be a bias as two people attending the interviews may have intimidated the interview partners. However, we avert this by selecting one main interviewer not to overcharge the interviewee.

## 6. Conclusion

Our research question was: "*How does the interplay of human and AI relate to persuasion success within a written message context?"* The analysis mainly shows that humans and AI complement each other. Certain artificial functionality provided by persuasive NLG (cf. Table 3, column 3) supports persuasive writers to achieve increased celerity (e.g., through automating the creation of logical coherence or conciseness) and also stronger alignment (e.g., through aligning formality or emotionality of recipient and persuader by quantifying both) which our experts recurrently related to as mirroring (IP1, IP2, IP6, cf. Voss & Raz 2016).